\documentclass{article}

    \PassOptionsToPackage{numbers, compress}{natbib}

    \usepackage[preprint]{neurips_2025}

\usepackage[utf8]{inputenc} 
\usepackage[T1]{fontenc}    
\usepackage{hyperref}       
\usepackage{url}            
\usepackage{booktabs}       
\usepackage{amsfonts}       
\usepackage{nicefrac}       
\usepackage{microtype}      
\usepackage{xcolor}         
\usepackage{multirow}

\usepackage{times}
\usepackage{epsfig}
\usepackage{graphicx}
\usepackage{amsmath}
\usepackage{amssymb}
\usepackage{booktabs}
\usepackage{pifont}    
\usepackage{subcaption}
\usepackage{url}
\usepackage{paralist}
\usepackage{wrapfig}
\usepackage{tabularx}

\title{\textit{GeoDrive}: 3D Geometry-Informed Driving World Model with Precise Action Control}

\author{
Anthony Chen$^{*1,2}$, Wenzhao Zheng$^{*3}$, Yida Wang$^{*2}$ \\
\textbf{Xueyang Zhang$^{2}$, Kun Zhan$^{2}$, Peng Jia$^{2}$, Kurt Keutzer$^{3}$, Shanghang Zhang$^{\dagger 1}$} \\
\\
$^1$State Key Laboratory of Multimedia Information Processing, \\
School of Computer Science, Peking University \\
$^2$Li Auto Inc. \\
$^3$UC Berkeley \\
\\
Code: \href{https://github.com/antonioo-c/GeoDrive}{\textcolor{magenta}{https://github.com/antonioo-c/GeoDrive}}
}

\begin{document}

\maketitle
\begin{figure}[h]
    \centering
    \vspace{-5mm}
    \includegraphics[width=0.98\linewidth]{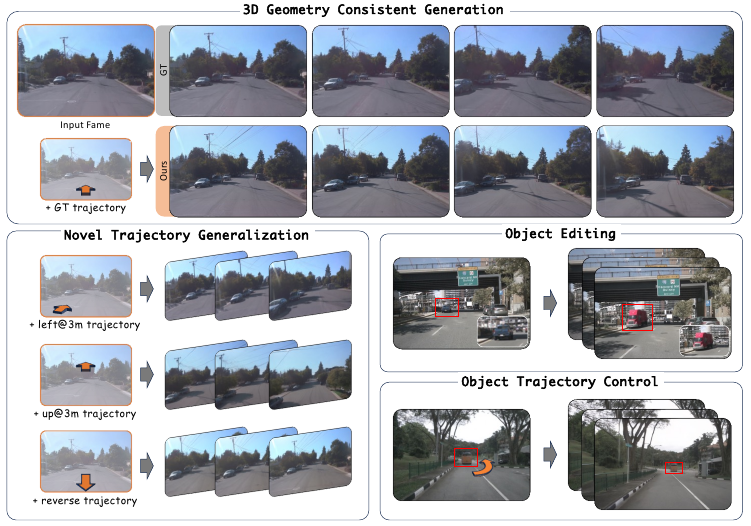}
    \vspace{-3mm}
    \small
    \caption{\textit{GeoDrive} enables precise trajectory following, correct novel view synthesis, and dynamic scene editing in autonomous driving scenarios. Our method integrates robust 3D conditions into driving world models, enhancing spatial understanding and action controllability.}
    

    \label{fig:teaser} 
\end{figure}

\renewcommand{\thefootnote}{\fnsymbol{footnote}} 
\footnotetext[1]{Equal Contribution.} 
\footnotetext[2]{Corresponding Author.} 

\begin{abstract}
  Recent advancements in world models have revolutionized dynamic environment simulation, allowing systems to foresee future states and assess potential actions. In autonomous driving, these capabilities help vehicles anticipate the behavior of other road users, perform risk-aware planning, accelerate training in simulation, and adapt to novel scenarios, thereby enhancing safety and reliability. Current approaches exhibit deficiencies in maintaining robust 3D geometric consistency or accumulating artifacts during occlusion handling, both critical for reliable safety assessment in autonomous navigation tasks.
  To address this, we introduce \textit{GeoDrive}, which explicitly integrates robust 3D geometry conditions into driving world models to enhance spatial understanding and action controllability.
  Specifically, we first extract a 3D representation from the input frame and then obtain its 2D rendering based on the user-specified ego-car trajectory. 
  To enable dynamic modeling, we propose a \textit{dynamic editing} module during training to enhance the renderings by editing the positions of the vehicles. 
  Extensive experiments demonstrate that our method significantly outperforms existing models in both action accuracy and 3D spatial awareness, leading to more realistic, adaptable, and reliable scene modeling for safer autonomous driving. Additionally, our model can generalize to novel trajectories and offers interactive scene editing capabilities, such as object editing and object trajectory control.

\end{abstract}

\section{Introduction}
Driving world models simulating 3D dynamic environments enable critical capabilities including trajectory-consistent view synthesis~\cite{yan2024street}, physics-compliant motion prediction~\cite{huang2023vad}, and safety-aware scenario reconstruction~\cite{yan2024street} and generation~\cite{gao2024vista,mao2024dreamdrive}.
Particularly, generative video models have emerged as effective tools for ego-motion forecasting and dynamic scene reconstruction~\cite{blattmann2023align,ho2022video,wang2023modelscope}. 
Their ability to synthesize trajectory-faithful visual sequences proves crucial for developing autonomous systems that anticipate environmental interactions while maintaining physical plausibility.

Despite these advancements, most existing methods lack sufficient 3D geometric awareness due to their reliance on 2D space optimization~\cite{gao2024vista}. 
This shortcoming results in structural incoherence across novel views and physically implausible object interactions~\cite{yang2024generalized, wang2023drivedreamer}, which is particularly detrimental for safety-critical tasks like collision avoidance in dense traffic. 
Also, existing methods usually depend on dense annotations (e.g., HD-map sequences and 3D bounding box tracks) for controllability~\cite{wang2023drivedreamer, li2023drivingdiffusion}, which only reproduce prescribed motions without understanding vehicle dynamics. 
A more flexible approach is to infer dynamic priors from single (or few) images while conditioning on the desired ego-trajectory. However, current methods that fine-tune on numerical camera parameters lack 3D geometry awareness, compromising their action controllability and consistency~\cite{gao2024vista, arai2024actbench, gaia1-2023}.
A reliable driving world model should satisfy three criteria: 1) rigid spatio-temporal coherence across static infrastructure and dynamic agents; 2) 3D controllability over ego-vehicle trajectories; and 3) kinematically constrained motion patterns for non-ego agents.

We achieve these demands through a hybrid neural-geometric framework that explicitly enforces 3D geometry consistency across generated sequences.
We first build a 3D structural prior from monocular input and then perform projective rendering along user-specified camera trajectories to generate geometrically grounded conditioning signals. 
We further employ cascaded video diffusion to refine these projections through 3D-attentive denoising, jointly optimizing photometric quality and geometric fidelity. 
For dynamic objects, we introduce a physics-guided editing module which transforms agent appearances under explicit motion constraints to ensure physically plausible interactions.
Our experiments demonstrate that GeoDrive significantly enhances the performance of controllable driving world models. 
Specifically, our method improves ego-car action controllability, reducing trajectory following errors by $42\%$ compared to the Vista~\cite{gao2024vista} model. 
Additionally, it achieves notable enhancements in video quality metrics, including LPIPS, PSNR, SSIM, FID, and FVD. Furthermore, our model generalizes effectively to novel view synthesis tasks, surpassing StreetGaussian in generated video quality. 
Beyond trajectory conditioning, GeoDrive offers interactive scene editing capabilities, such as dynamic object insertion, replacement, and motion control. Additionally, by integrating real-time visual input with predictive modeling, we enhance the decision-making process of vision-language models, providing an interactive simulation environment that enables safer and more effective trajectory planning.

\section{Related Work}

\textbf{Driving World Models.}
World models have become a cornerstone for enabling intelligent agents to anticipate and act in complex, dynamic environments, with autonomous driving presenting unique challenges due to its large field of view, highly dynamic scenes, and the need for robust generalization. Recent research has explored a variety of generative frameworks for future prediction, leveraging representations such as point clouds~\cite{fan2019pointrnn, lu2021monet, khurana2022differentiable, zhangcopilot4d, huang2025neural, zhang2024nerf, wu2023fast, huang2023neural, yang2024visual}, occupancy grids~\cite{liu2024lidar, bogdoll2023muvo, ma2024cam4docc, zheng2025occworld, gu2024dome, min2024driveworld, zuo2024gaussianworld, wei2024occllama, wang2024occsora, yan2024renderworld}, and images~\cite{wang2023drivedreamer, zhao2024drivedreamer, zhao2024drivedreamer4d, ni2024recondreamer, wang2024worlddreamer, doe, chi2024evaembodiedworldmodel}.
Point-cloud-based methods leverage the detailed geometric information captured by LiDAR to predict future states and enable precise modeling of spatial geometry and dynamic interactions~\cite{zhangcopilot4d, huang2025neural, zhang2024nerf}.
Occupancy-grid-based methods further discretize the environment into voxel grids for more fine-grained and geometrically consistent modeling of scene evolutions~\cite{ma2024cam4docc, zheng2025occworld, gu2024dome, min2024driveworld, zuo2024gaussianworld, wei2024occllama, wang2024occsora, yan2024renderworld}.

Image-based world models stand out for holding more promise for scaling up due to sensor flexibility and data accessibility~\cite{wang2023drivedreamer, zhao2024drivedreamer, zhao2024drivedreamer4d, ni2024recondreamer, wang2024worlddreamer, gao2024cardreamer, li2023drivingdiffusion, ma2024unleashing}.
They usually leverage powerful generative models to capture the complex visual dynamics of real-world environments, making them particularly valuable for perception and planning tasks~\cite{hu2024drivingworld, hu2023gaia, sreeram2024probing, wu2024holodrive, zhang2024bevworld, swerdlow2024street, wang2024drivingdojo, yang2024drivearena, yan2024drivingsphere, garg2024imagine, gao2024vista, zhou2024simgen, li2024enhancing, popov2024mitigating}. Although existing generative models (e.g., DriveDreamer~\cite{wang2023drivedreamer} and DrivingDiffusion~\cite{li2023drivingdiffusion}) achieve accurate scene control by conditioning on dense annotations (e.g., HD‑map sequences and long tracks of 3D bounding boxes), they can only reproduce prescribed motions without truly understanding vehicle dynamics. 
A more flexible alternative is to infer dynamic priors directly from a single (or a few) images while simultaneously conditioning on the desired ego‑trajectory.
Recent systems such as Vista~\cite{gao2024vista}, Terra~\cite{arai2024actbench}, and GAIA 1\&2~\cite{gaia1-2023,russell2025gaia2controllablemultiviewgenerative} enable action-conditioned generation by injecting raw numerical control vectors directly into the generative backbone. 
However, since control vectors are not explicitly grounded in the visual latent space, the resulting action signals are weak and often lead to unstable control, demanding larger training datasets for convergence~\cite{gao2024vista, gaia1-2023}. 
Differently, our approach renders action commands as visual conditioning inputs that are naturally aligned with the generative latent space, which delivers a substantially stronger control signal and yields significantly more stable and reliable generation results.

\textbf{Conditional Video Generation.}
Generative diffusion models have evolved from text‑to‑image systems into fully multimodal engines that can synthesize entire video sequences on demand. Throughout this progression, the research focus has steadily shifted toward conditional generation—giving users explicit levers to guide the output. Milestones such as~ControlNet~\cite{zhang2023adding}, T2I‑Adapter~\cite{mou2024t2i}, and GLIGEN~\cite{li2023gligen} first embedded conditioning signals into text‑to‑image pipelines; follow‑up studies extended the idea to video, allowing control with RGB key‑frames~\cite{blattmann2023svd,xing2023dynamicrafter,xing2024tooncrafter}, depth maps~\cite{xing2024make,esser2023structure}, object trajectories~\cite{yin2023dragnuwa,niu2024mofa}, or semantic masks~\cite{peruzzo2024vase, chen2024trainingfreeregionalpromptingdiffusion}. Yet steering the 6‑DoF camera path remains difficult. Coarse LoRA‑based motion classes~\cite{guo2023animatediff,blattmann2023svd}, numeric‑matrix conditioning~\cite{wang2023motionctrl}, depth‑warping schemes~\cite{muller2024multidiff}, and Plücker‑coordinate encodings~\cite{xu2024camco,he2024cameractrl} each fall short—either through imprecise control, limited domain coverage, or an indirect mapping from numbers to pixels.

Planners and safety modules require frame-level accuracy, so generators such as DriveDreamer \cite{wang2023drivedreamer} and DrivingDiffusion~\cite{li2023drivingdiffusion} lean on dense HD-map sequences and long 3D box tracks to lock the scene to a predefined route. Other systems like Vista~\cite{gao2024vista}, GAIA 1\&2~\cite{gaia1-2023,russell2025gaia2controllablemultiviewgenerative}, add control vectors directly into the backbone features, but the mismatch between numerical commands and visual features weakens the signal, slows optimisation, and often produces drift. In this work, we propose to use explicit visual conditions for accurate ego trajectory control.

\begin{figure*}[t]
  \centering
   \includegraphics[width=0.95\linewidth]{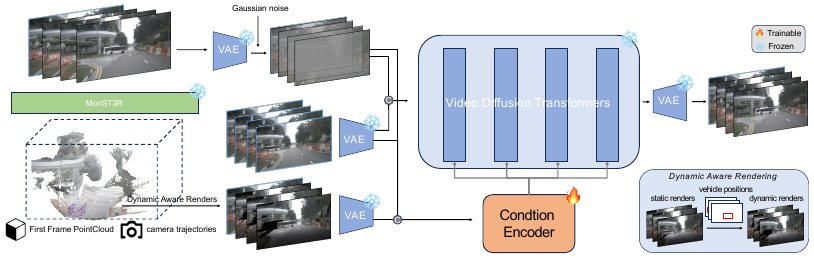}
   \vspace{-3mm}
   \caption{\textbf{Overview of our training pipeline.} We use a pretrained dense stereo model to obtain 3D point clouds and camera trajectories. A dynamic video is rendered from the first-frame point cloud using our \textit{dynamic editing} technique. The noisy latent representation and rendered video are encoded via a VAE and concatenated as input for our \textit{condition encoder}, modulating the DiT model's features. The DiT then generates photorealistic video that accurately follows the specified action conditions.}
   \vspace{-3mm}
   \label{fig:pipeline}
\end{figure*}

\section{Methodology}
\label{sec:methodology}

Given an initial reference image \( I_{0} \in \mathbb{R}^{H \times W \times 3} \) and ego-vehicle trajectory \( \{C_t\}_{t=1}^L \), our framework synthesizes realistic future frames that follow the input trajectory. We leverage 3D geometric information from the reference image to guide world modeling. First, we reconstruct a 3D representation (Sec.~\ref{sec:3d_reconstruction}), then render video sequences along user-specified trajectories with dynamic object handling (Sec.~\ref{sec:projection}). The rendered video provides geometric guidance for generating spatio-temporally consistent videos that follow the input trajectory (Sec.~\ref{sec:diffusion}). See Figure~\ref{fig:pipeline} for illustration.

\subsection{Extracting 3D Representations from Reference Image}
\label{sec:3d_reconstruction}
To utilize 3D information for 3D-consistent generation, we first construct a 3D representation from the single input image \( I_0 \). We employ MonST3R~\cite{zhang2024monst3r}, an off-the-shelf dense stereo model that simultaneously predicts 3D geometry and camera poses, aligning with our training paradigm. During inference, we duplicate the reference image to satisfy MonST3R's cross-view matching requirements.

Given RGB frames \(\{I_t\}_{t=0}^T\), MonST3R predicts per-pixel 3D coordinates \(\{O_t\}_{t=0}^T\) and confidence scores \(\{D_t\}_{t=0}^T\) via cross‐view feature matching across frames
\begin{equation}
  \{O_t\}_{t=0}^T, \{D_t\}_{t=0}^T = \mathrm{MonST3R}\bigl(\{I_t\}_{t=0}^T\bigr),
  \label{eq:monst3r_output}
\end{equation}
where \(\mathbf{O}_t^{i,j}\in\mathbb{R}^3\) denotes the metric‐space position of pixel \((i,j)\) in the \(t^{th}\) reference frame, and \(\mathbf{D}_t^{i,j}\in[0,1]\) measures reconstruction reliability. By thresholding \(\mathbf{D}_0\) at \(\tau\) (typically \(\tau=0.65\)), the colored point cloud for the \(t^{th}\) reference frame yields as
\begin{equation}
  \mathcal{P}_t = \bigl\{(\mathbf{O}_t^{i,j},\,I_t^{i,j}) \mid \mathbf{D}_t^{i,j}>\tau\bigr\}.
  \label{eq:point_cloud}
\end{equation}
To counteract the imbalance between valid and invalid matches across the sequence, the confidence map \(\mathbf{D}_0\) is trained with a focal loss. Further, to disentangle static scene geometry from moving objects, MonST3R employs a transformer-based decoupler. This module processes the initial features of the reference frame (enriched by cross-view context) and separates them into static and dynamic components. The decoupler uses learnable prompt tokens to split attention maps: static tokens attend to large planar surfaces, and dynamic tokens attend to compact, motion-rich regions. By excluding dynamic correspondences, we obtain a robust camera pose estimate
\begin{equation}
  \hat{C}_t = \arg\min_{C_t}\sum_{(i,j)\in\mathbf{F}^{\mathrm{static}}}
    \bigl\|\pi(C_t\,\mathbf{O}_t^{i,j})-\mathbf{p}_t^{i,j}\bigr\|_2^2,
  \label{eq:pose_estimation}
\end{equation}
where \(\pi\) denotes the perspective projection operator, and only static feature matches are used. Compared to conventional structure‑from‑motion \cite{dust3r}, this strategy reduces pose error by 38\% in dynamic urban scenes \cite{zhang2024monst3r}. The resulting point cloud \(\mathcal{P}_0\) then serves as our geometric scaffold.

\begin{figure*} [t] 
  \centering
     \includegraphics[width=0.95\linewidth]{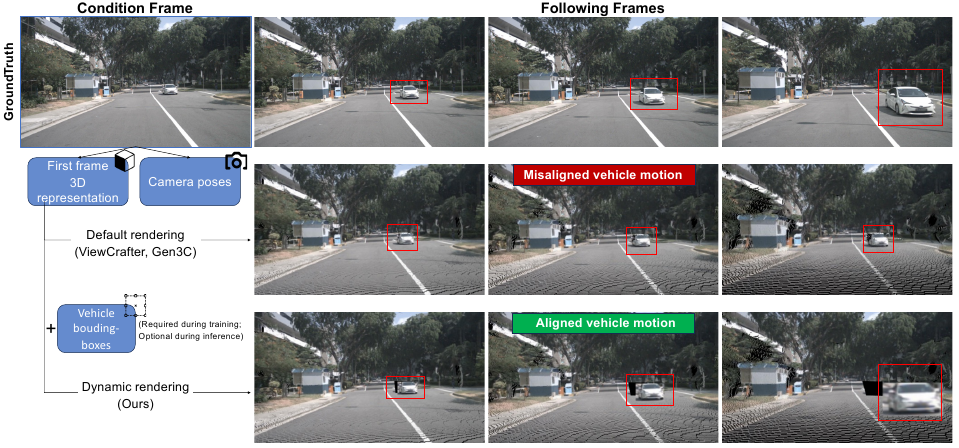}
   \vspace{-3mm}
  \caption{\textbf{Illustration of \textit{dynamic edit} design.} Compared with default rendering, it effectively reduces disparity between static rendering and dynamic real-world scenarios.}
   \vspace{-1mm}
   \label{fig:vis_dynamicedit}
\end{figure*}

\subsection{Rendering 3D Videos with Dynamic Editing} 
\label{sec:projection}
To achieve precise input trajectory following, our model renders a video that serves as a visual guide for the generation process. We project the reference point cloud $\mathcal{P}_0$ through each user-provided camera configuration $C_t = (R_t, T_t, f_t)$ using standard projective geometry techniques. Each 3D point $\mathbf{P}_i^w \in \mathcal{P}_0$ undergoes a rigid transformation into the camera coordinate system
$\mathbf{P}_i^c = R_t \mathbf{P}_i^w + T_t$,
followed by perspective projection using the camera's intrinsic matrix $K_t$,
yielding image coordinates $\mathbf{p}_i = \left( \frac{f_t \mathbf{P}_i^{c_x}}{\mathbf{P}_i^{c_z}} + \frac{W}{2}, \frac{f_t \mathbf{P}_i^{c_y}}{\mathbf{P}_i^{c_z}} + \frac{H}{2} \right).$
We only consider valid projections within a depth range of $\mathbf{P}_i^{c_z} \in [0.1, 100.0]$ meters and use z-buffering to handle occlusions, ultimately producing the rendered view $\tilde{I}_t$ for each camera position.

\textbf{Limitations of Static Rendering.} 
Since we utilize only the first frame point cloud, the rendered scene remains static throughout the sequence. This creates a significant discrepancy with real-world autonomous driving contexts, where vehicles and other dynamic objects are in constant motion. The static nature of our rendering fails to capture the dynamic essence that distinguishes autonomous driving datasets from traditional static scenes.

\begin{table*}[t] \small
\centering
\caption{\textbf{Quantitative results of generation quality and action fidelity on NuScenes~\cite{caesar2020nuscenes} validation subset.} We outperform baseline methods on every metric while requiring much less training data.}
\vspace{-3mm}
\resizebox{\linewidth}{!}
{
\begin{tabular}{l|c|ccccc|cc}
\toprule
\multirow{2}{*}{\textbf{Method}} & \multirow{2}{*}{\textbf{Data Scale}} & \multicolumn{5}{c}{\textbf{Prediction Quality}} & \multicolumn{2}{c}{\textbf{Action Fidelity}} \\
\cmidrule(lr){3-7} \cmidrule(lr){8-9}
& & \textbf{LPIPS $\downarrow$} & \textbf{PSNR $\uparrow$} & \textbf{SSIM $\uparrow$} & \textbf{FID $\downarrow$} & \textbf{FVD $\downarrow$} & \textbf{ADE$_{^{\times 10^2}}$ $\downarrow$} & \textbf{FDE$_{^{\times 10^2}}$ $\downarrow$} \\ 
\midrule
Vista~\cite{gao2024vista}  & 1740h        & 0.351          & 20.086        & 0.621         & 8.35        & 163.7 & 2.77 &  5.28 \\
Terra~\cite{arai2024actbench}   & 1740h        & 0.455              & 18.42               & 0.553               & 26.73             & 787.54   & 5.57 &  11.8  \\ 
\midrule
GeoDrive (Ours) & 5h & \textbf{0.303} & \textbf{21.979} & \textbf{0.6535} & \textbf{7.17} & \textbf{85.22} & \textbf{1.62} & \textbf{3.1} \\

\bottomrule
\end{tabular}
}
\vspace{-1mm}
\label{tab:quanti}
\end{table*}

\begin{figure*} [t]
  \centering
   \includegraphics[width=0.99\linewidth]{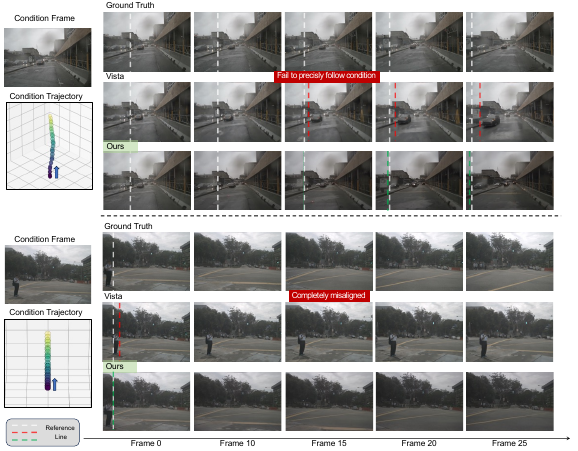}
   \vspace{-3mm}
  \caption{\textbf{Qualitative comparison of action fidelity under the same conditional frame and action control.} Our model precisely follows desired trajectory, while Vista~\cite{gao2024vista} produce misaligned results.}
  \label{fig:trajectory}
   \vspace{-1mm}
\end{figure*}

\textbf{Dynamic Editing.} To address this limitation, we propose \textit{dynamic editing} to produce renderings \(R\) with static backgrounds and moving vehicles. Specifically, when users provide a sequence of 2D bounding box information for moving vehicles in the scene, we dynamically adjust their positions to create the illusion of motion in the rendering. This approach not only guides the ego-vehicle's trajectory during the generation process but also directs the movement of other vehicles in the scene. Fig.~\ref{fig:vis_dynamicedit} provides an illustration of this process. Such a design significantly reduces the disparity between static rendering and dynamic real-world scenarios while enabling flexible control over other vehicles—a capability that existing methods like Vista~\cite{gao2024vista} and GAIA~\cite{gaia1-2023} fail to achieve.

\subsection{Dual-Branch Control for Spatio-Temporal Consistency}  
\label{sec:diffusion}  
While the point cloud-based rendering accurately preserves geometric relationships between views, it suffers from several visual quality issues. The rendered views often contain substantial occlusions, missing areas due to limited sensor coverage, and reduced visual fidelity compared to real camera images. To enhance the quality, we adapt a latent video diffusion model~\cite{video_ldm} to refine projected views while preserving 3D structural fidelity through specialized conditioning.

\begin{figure*}[t] 
  \centering
  \includegraphics[width=0.99\linewidth]{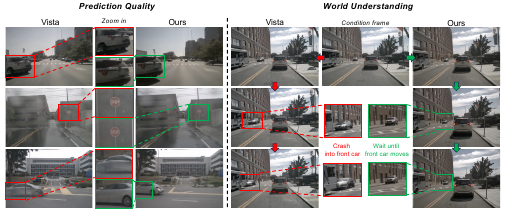}
  \vspace{-3mm}
  \caption{\textbf{Qualitative Comparisons:} Left - Enhanced visual fidelity in our predictions; Right - Superior scene dynamics understanding.}
  \label{fig:quality}
  \vspace{-7mm}
\end{figure*}

Building on this, we further refine the integration of contextual features into a pre-trained diffusion transformer (DiT), drawing inspiration from the methodology introduced by VideoPainter~\cite{bian2025videopainteranylengthvideoinpainting}. However, we introduce key distinctions tailored to our specific needs. We employ dynamic renderings to capture temporal and contextual nuances, providing a more adaptive representation for the generation process. Let $\delta_{\phi}(z_t, t, C)$ represent the feature output at layer $i$ of our modified DiT backbone $\delta_{\phi}$, where $z_R$ denotes the dynamic renderings latent via VAE encoder $\mathcal{E}$ and $z_t$ is the noisy latent at timestep $t$. 

These renderings are processed through a lightweight condition encoder, which extracts essential background cues without duplicating extensive portions of the backbone architecture. The integration of features from the condition encoder into the frozen DiT is formulated as follows:
\begin{equation}
\delta_{\phi}(z_t, t, C)_i = \delta_{\phi}(z_t, t, C)_i + \mathcal{W} \left( \gamma_{\phi}^{enc}([z_t, z_R], t)_{i // \frac{M}{2}} \right),
\label{eq:integration}
\end{equation}
where $\gamma_{\phi}^{enc}$ denotes the condition encoder processing the concatenated input of noisy latent $z_t$ and renderings latent $z_R$, with $M$ representing the total number of layers in the DiT backbone. $\mathcal{W}$ is a learnable linear transformation initialized to zero to prevent noise collapse in early training. The extracted features are selectively fused into the frozen DiT in a structured manner, ensuring that only relevant contextual information guides the generation process. The final video sequence is decoded via the frozen VAE decoder $\mathcal{D}$ as $\hat{I}_t = \mathcal{D}(z_t^{(0)})$. 

By limiting training to condition encoder $g\phi$ alone (6\% of total parameters), we maintain the pre-trained model's photorealism and gain precise camera control. Temporal coherence arises naturally from the video transformer's dynamics modeling and the geometric consistency of $\{\tilde{I}_t\}$ features across frames, enabling trajectory-faithful video synthesis.

\begin{figure*} [t]
  \centering
    \includegraphics[width=0.95\linewidth]{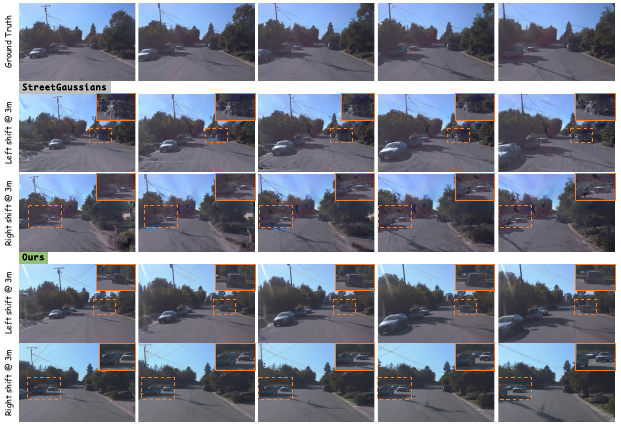}
  \vspace{-3mm}
  \caption{\textbf{Qualitative comparison on novel-view synthesis on Waymo validation subset.} Our model generates sharp results for deviated trajectories in a zero-shot manner, whereas the reconstruction-based method StreetGaussian~\cite{yan2024street} produces significant artifacts.}
  \label{fig:noveltraj}
  \vspace{-3mm}
\end{figure*}

\section{Experiments and Applications}
\label{sec:experiments}


\subsection{Experimental Settings}
\label{sec:implementation}

\textbf{Training Configuration.} We train exclusively on nuScenes~\cite{caesar2020nuscenes}, processing each clip through MonST3R to obtain metric-scale 3D reconstructions and camera trajectories. 
3D reconstructions of the initial frame $\mathcal{P}_0$ undergoes projective rendering along estimated trajectories via a differentiable rasterizer, where \textit{Dynamic Editing} leverages 2D bounding box annotations to edit vehicle positions. 
We curate 25,109 video-condition pairs for training. We freeze the base diffusion model (CogVideo-5B-I2V~\cite{hong2022cogvideo}) while training the condition encoder for 28,000 steps at learning rate 
$1\times10^{-5}$ for 4 days. More details on training, inference and model configuration are in Appendix~\ref{sec:append_exp_details}.




\subsection{Trajectory Following}
\label{sec:result_trajectory}

\begin{wraptable}{r}{0.5\textwidth} 
   \small
   \centering
   \scriptsize
   \vspace{-5mm}
   \caption{\textbf{Quantitative results of generation quality on NuScenes validation fullset.}}
   \label{tab:quanti_full}
   \begin{tabular}{lccc}
     \toprule
     \textbf{Models} & \textbf{Data Scale} & \textbf{FID} $\downarrow$ & \textbf{FVD} $\downarrow$ \\
     \midrule
     DriveGAN~\cite{kim2021drivegan} & 160h & 73.4 & 502.3 \\
     DriveDreamer~\cite{wang2023drivedreamer} & 5h & 14.9 & 340.8 \\
     DriveDreamer-2~\cite{zhao2024drivedreamer} & 5h & 25.0 & 105.1  \\
     WoVoGen~\cite{lu2025wovogen} & 5h & 27.6 & 417.7  \\
     Drive-WM~\cite{min2024driveworld} & 5h & 15.8 & 122.7  \\
     GenAD~\cite{yang2024generalized} & 1740h & 15.4 & 184.0  \\
     \midrule
     Vista~\cite{gao2024vista} & 1740h & 6.6 & 167.7  \\
     GEM~\cite{hassan2024gem} & 4000h & 10.5 & 158.5 \\
     GeoDrive (Ours) & 5h & \textbf{4.1} & \textbf{61.6} \\
     \bottomrule
   \end{tabular}
   \vspace{-5mm}
\end{wraptable}

\textbf{Benchmark and Baselines.} 
We compare GeoDrive to two most-relevant baselines that condition on single image and ego action (Vista~\cite{gao2024vista}, Terra~\cite{arai2024actbench}), as well as several other driving world models.
We adhere to Vista's protocol by computing trajectory from sensor and calibration data that spans the 25-frame clip, as their condition input. We estimate our condition camera poses by running MonST3R on GT video. While we condition on different modalities, the trajectories for all methods are extracted from the same ground-truth video clip, ensuring aligned action conditions. We evaluate all methods on NuScenes validation set. For trajectory control precision evaluation, we sample a subset of 1087 videos with balanced driving trajectories. Visual quality is quantified through PSNR, SSIM~\cite{wang2004image}, LPIPS~\cite{Johnson2016Perceptual}, FID~\cite{heusel2017gans}, and FVD~\cite{unterthiner18}. While trajectory fidelity metrics employ Average Displacement Error (ADE) and Final Displacement Error (FDE).

\begin{figure} [t]
\centering
\includegraphics[width=\linewidth]{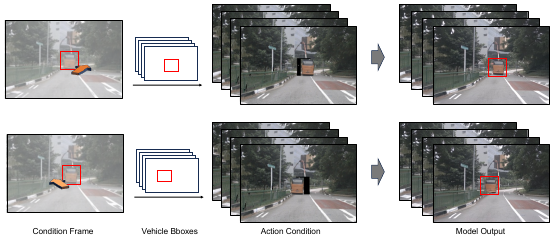}
  \vspace{-6mm}
\caption{\textbf{Qualitative Results on Vehicle Manipulation.} Our approach allows for the manipulation of vehicle movement directions within a scene by specifying different bounding boxes.}
\vspace{-1mm}
\label{fig:manipulation}
\end{figure}%

\textbf{Quantitative Results.}
The quantitative results on nuScenes validation subset are presented in Table~\ref{tab:quanti}. GeoDrive outperforms baselines in all metrics. Specifically, our trajectory-following ability is significantly better than the baselines, yet requiring 99.7\% less data, showing the effectiveness of our method. In Table~\ref{tab:quanti_full}, GeoDrive outperforms all baselines on FID \& FVD results.

\textbf{Qualitative Results.}
As shown in Figure~\ref{fig:trajectory}, our method produce agents adhering more precisely to the specified paths compared to baseline methods, which often deviate from the intended trajectory.
In addition, our generated videos exhibit a better understanding of the driving environment. As illustrated in Figure~\ref{fig:quality}, our results are noticeably sharper and contain fewer artifacts. Furthermore, our model demonstrates a stronger grasp of scene dynamics compared to baseline methods. For example, in Figure~\ref{fig:quality}, our model correctly anticipates that the car behind should wait for the bus ahead to move, whereas Vista erroneously accelerates the car forward, resulting in a collision.
Moreover, we examine model performance on reverse trajectory, which is not included in the training data. Demonstrated in Figure~\ref{fig:qual_reverse} in Appendix~\ref{sec:append_results}, GeoDrive successfully generalizes to the unseen trajectory, while Vista fails to perform. Moreover, thanks to our 3D geometric condition, the building structure remains exactly the same across generation. 

\begin{table}[t] \small
  \centering
  \begin{minipage}{0.53\textwidth}
    \caption{\textbf{Quantitative results for NVS on Waymo validation subset.} GeoDrive is trained solely on the NuScenes dataset, yet it can generalize to Waymo scenes in a zero-shot, feed-forward manner.}
    \begin{tabularx}{\textwidth}{l|X|X|X|X}
      \toprule
      \multirow{2}{*}{\textbf{Method}} & \multicolumn{2}{c}{\textbf{Left@3m}} & \multicolumn{2}{c}{\textbf{Right@3m}} \\
      \cmidrule(lr){2-3} \cmidrule(lr){4-5}
      & \textbf{FID $\downarrow$} & \textbf{FVD $\downarrow$} & \textbf{FID $\downarrow$} & \textbf{FVD $\downarrow$} \\
      \midrule
      StreetGS~\cite{yan2024street} & \textbf{63.84} & 1438.89 & 69.55 & 1526.62 \\
      \midrule
      Ours & 67.13 & \textbf{1245.23} & \textbf{65.67} & \textbf{1422.63} \\
      \bottomrule
    \end{tabularx}
    \label{tab:novel_view}
  \end{minipage}\hfill
  \begin{minipage}{0.44\textwidth}
    \caption{\textbf{Ablation Studies on NuScenes validation subset.} D.E. represents dynamic editing. \textit{w/o} dual-branch means we adjust input channels to adapt renders and finetune backbone.}
    \begin{tabularx}{\textwidth}{l|X|X|X}
        \toprule
        \textbf{Method} & \textbf{FID} & \textbf{FVD} & \textbf{ADE$_{^{\times 10^2}}$} \\
        \midrule
         \textit{w/o} \textit{D.E.} & 7.01 & 88.68 & 3.68 \\
         \textit{w/o} dual-branch & 8.83 & 74.76 & 3.45 \\
         GeoDrive & 7.17 & 85.22 & 1.62 \\
        \bottomrule
    \end{tabularx}
    \label{tab:ablation}
  \end{minipage}
  \vspace{-4mm}
\end{table}

\subsection{Novel View Synthesis}
\label{sec:result_novelview}
    
\textbf{Benchmark and Baseline.}
We compare GeoDrive to scene reconstruction method, StreetGaussians \cite{yan2024street}.
We evaluate on the Waymo validation set and filter out 5 scenes for testing. The novel trajectory is created by horizontally shifting from the original trajectory of the frontal camera. We assess generation quality using FID and FVD since there
is no ground truth for novel trajectories.

\textbf{Quantitative Results.}
As demonstrated in Table~\ref{tab:novel_view}, our method achieves lower (and thus better) FID \& FVD scores compared to the reconstruction-based baselines. This is due to the difficulty reconstruction methods face in recovering scene structures from the sparsely observed views.

\textbf{Qualitative Results.}
Figure~\ref{fig:noveltraj} illustrates that while StreetGaussians~\cite{yan2024street} generates projectively correct renderings along given trajectories, it exhibits severe geometric distortions under viewpoint shifts. Such degrades reveal fundamental limitations in 3D scene reconstruction from sparse observational data, particularly the inability to resolve occlusion boundaries and low-textured supervisions. GeoDrive maintains trajectory adherence while preserving photorealistic rendering.

\begin{figure} [t]
\centering
\includegraphics[width=\linewidth]{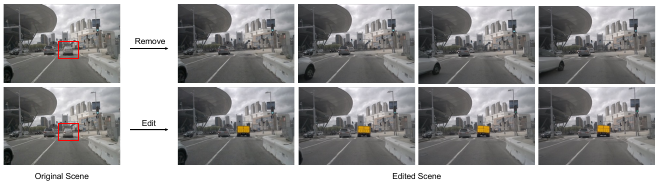}
\vspace{-6mm}
\caption{\textbf{Qualitative Results on Scene Editing.} Our approach enables the removal or replacement of vehicles within a scene, allowing for the prediction of seamless future scenarios.}
  \vspace{-5mm}
\label{fig:edit}
\end{figure}

\begin{figure} [t]
    \centering
    \includegraphics[width=0.99\linewidth]{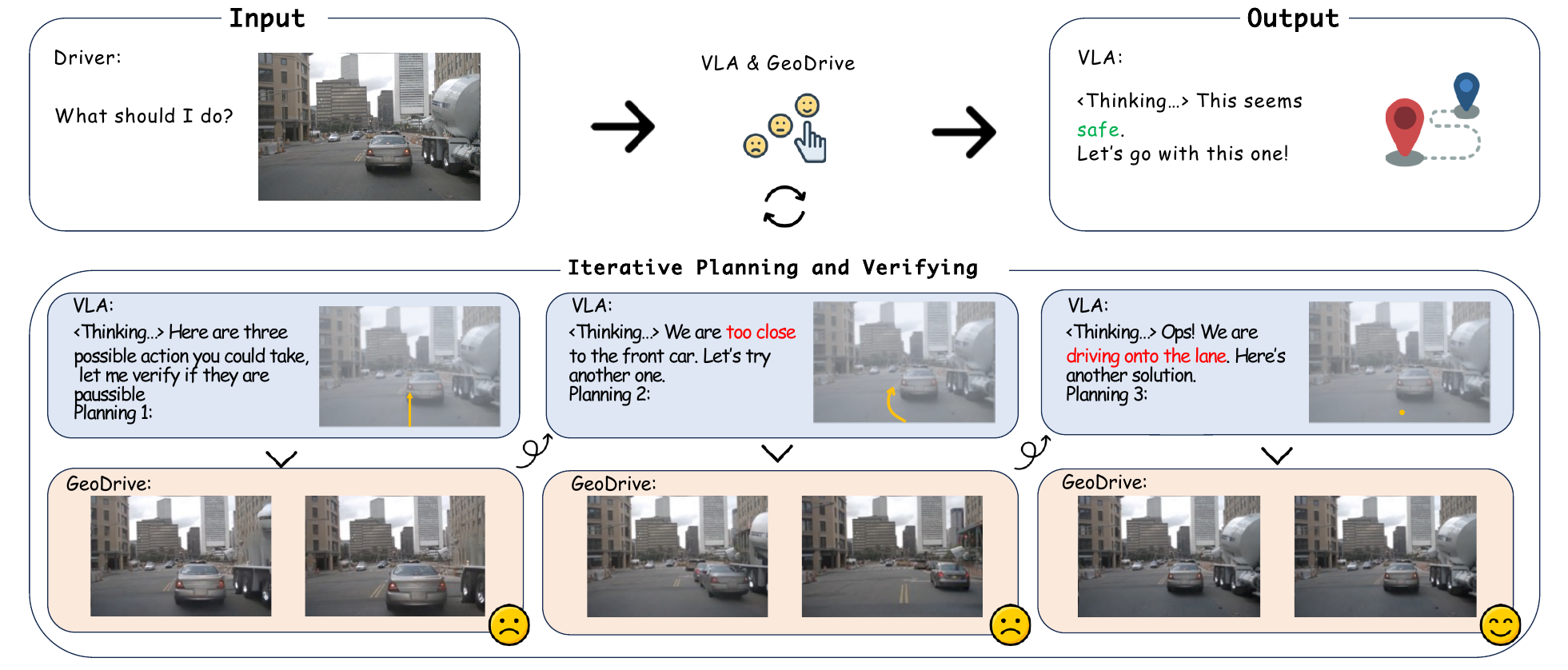}
  \vspace{-3mm}
    \caption{\textbf{Illustration of application to VLA planning.} By simulating each possible planned trajectory, we can assist the VLA model in refining its decisions until it reaches the optimal decision.}
  \vspace{-1mm}
     \label{fig:VLA}
\end{figure}

\subsection{Applications}
\label{sec:application}
In this section, we demonstrate the versatility and practical impact of GeoDrive across several key applications in autonomous driving. It not only supports advanced scene editing, such as object editing and object trajectory control, but also it can serve as an interactive environment that assists high-level planning models, such as the VLA, in making safer and more informed decisions.

\textbf{Object Trajectory Control.}
As illustrated in Figure~\ref{fig:manipulation}, given extra condition input (i.e., a sequence of 2D bounding boxes) for other vehicles, we can apply \textit{dynamic augmentation} (Sec.~\ref{sec:projection}) and thus control the trajectory of vehicles appearing in the condition frame.

\textbf{Object Editing.}
GeoDrive supports intuitive and powerful scene editing capabilities, including modifying existing objects or removing unwanted elements from the environment. This is achieved by using off-the-shelf image editing models~\cite{flux2024} on the condition frame. As shown in Fig.~\ref{fig:edit}, our method offers significant flexibility in manipulating driving scenes. This level of control is invaluable for generating targeted training data, analyzing the influence of specific objects or configurations, and creating diverse scenarios from a limited amount of captured data.

\textbf{Assistance for VLA planning.}
GeoDrive enhances the decision-making of VLA (Visual Language Action) models by providing an interactive simulation environment for evaluating driving actions (~\cite{zhou2025opendrivevlaendtoendautonomousdriving, tian2024drivevlmconvergenceautonomousdriving, zhao2025sce2drivexgeneralizedmllmframework, guo2025vdtautoendtoendautonomousdriving}). As illustrated in Figure~\ref{fig:VLA}, it integrates real-time visual input with predictive modeling, allowing the VLA system to simulate outcomes of planned trajectories. This helps foresee hazards, such as proximity to other vehicles or lane deviations, and assess action safety. Consequently, the VLA model refines its decisions by selecting actions predicted to be safe and effective, which ensures driving decisions are contextually appropriate and prioritize safety.


\subsection{Ablation Studies}
\label{sec:ablation}

\textbf{Impact of Dynamic Editing.}
To quantify the impact of our dynamic editing strategy, we report the performance of our model trained with and without this component on key evaluation metrics as shown in Table~\ref{tab:ablation}.

\textbf{Impact of Dual-branch Architecture.}
We further investigate the effectiveness of adopting a dual-branch architecture compared with single-branch architecture. We evaluate the performance of different architectures on key evaluation metrics. The results are shown in Table~\ref{tab:ablation}.

\section{Conclusions and Limitations}
\label{sec:conclusion}

We presented GeoDrive, a video diffusion world model for autonomous driving, enhancing action controllability and spatial accuracy via explicit meter-level trajectory control and direct visual conditioning. Our method reconstructs 3D scenes, renders along desired trajectories, and refines the output with video diffusion. Evaluations demonstrate superior performance over existing models in visual realism and action adherence, enabling applications like non-ego view generation and scene editing, thus setting a new benchmark. However, our performance depends on the accuracy of depth and pose estimation from MonST3R, and the reliance solely on image and trajectory input for world prediction poses a challenge. Future work will explore incorporating text conditions and VLA understanding to further improve realism and consistency.

{\small
\bibliographystyle{plain}
\bibliography{egbib}
}

 \newpage
\appendix

\renewcommand\thesection{\Alph{section}}

\section*{\Large Appendix}

\setcounter{section}{0}


\section{Preliminary Details}
\subsection{Video Diffusion Models}
\label{subsec:prelimiary}
A diffusion model~\cite{song2021denoising} is built from two stochastic chains: a \emph{forward} (noising) process $q$ and a \emph{reverse} (denoising) process $p_{\theta}$.  
Starting with a clean sample $\mathbf{x}_0\!\sim\!q_0(\mathbf{x}_0)$, the forward process incrementally injects Gaussian noise, producing
\(
\mathbf{x}_t=\alpha_t\mathbf{x}_0+\sigma_t\boldsymbol{\epsilon},
\;\boldsymbol{\epsilon}\sim\mathcal{N}(\mathbf{0},\mathbf{I}),
\)
where the schedule satisfies $\alpha_t^{2}+\sigma_t^{2}=1$.  
The reverse process seeks to remove this noise using a neural predictor $\boldsymbol{\epsilon}_{\theta}$, trained by minimizing
\begin{equation}
\mathcal{L}_{\text{DM}}
=\mathbb{E}_{t\sim\mathcal{U}(0,1),\,
           \boldsymbol{\epsilon}\sim\mathcal{N}(\mathbf{0},\mathbf{I})}
\!\left[
  \bigl\|
    \boldsymbol{\epsilon}_{\theta}(\mathbf{x}_t,t)
    -
    \boldsymbol{\epsilon}
  \bigr\|_2^{2}
\right].
\end{equation}

To ease the heavy computation of pixel‑space generation, latent diffusion models (LDMs)~\cite{metzer2022latent} compress each RGB video
$\mathbf{x}\!\in\!\mathbb{R}^{L\times3\times H\times W}$
into a lower‑dimensional tensor
$\mathbf{z}=\mathcal{E}(\mathbf{x})\!\in\!\mathbb{R}^{L\times C\times h\times w}$
via a frozen VAE encoder $\mathcal{E}$.  
Both the forward and reverse chains operate in this latent space, after which a decoder $\hat{\mathbf{x}}=\mathcal{D}(\mathbf{z})$ reconstructs the final video.

In this paper, we adopt the pretrained CogVideo-I2V\cite{yang2024cogvideox} model as our backbone. Its ability to animate a single reference image aligns naturally with our objective of predicting future scenarios based on a single input.

\section{Experimental Details}
\label{sec:append_exp_details}

\subsection{Model Configuration}
GeoDrive is built upon a pre-trained image-to-Video Diffusion
Transformer CogVideo-5B-I2V~\cite{hong2022cogvideo}. Our lightweight condition encoder clones only the first two layers of
pre-trained DiT, taking up only 6\% of the backbone parameters.

\subsection{Training Details}
We train GeoDrive at $480 \times 720$ resolution, with learning rate $1 \times 10^{-5}$ and batch size 1 with AdamW~\cite{kingma2014adam} optimizer. During training, only the parameters of our condition encoder are optimized for 28000 steps. The model is trained with 8xA100 80GB GPUs for 4 days.

\subsection{Inference Details}
During inference, given a condition frame input and a trajectory (i.e. camera poses), we first build a 3D representation from the single image. In order to utilize MonST3R~\cite{zhang2024monst3r}, we duplicate the input image and get a video input for MonST3R. Next, we render a sequence of images along user-provided camera poses. GeoDrive can optionaly accept object bounding boxes to achieve control over object's trajectory in the scene, via dynamic editing module (Section~\ref{sec:projection}.

For \textbf{Trajectory Following} experiments (Section~\ref{sec:result_trajectory}), we use the first frame of each video as condition frame, and we estimate trajectory with MonST3R from the video. For fair comparision with baseline methods, We do not include object control as the baseline methods do not accept object bounding box information.

For \textbf{Novel View Synthesis} experiments (Section~\ref{sec:result_novelview}), we use the first frame of each video as condition frame, and we estimate the original trajectory with MonST3R from the video. Next, we align the original trajectory with the depth scale from the Lidar point cloud. Then we shift the trajectory left, right and up to obtain novel trajectory input for the experiment.

\subsection{Benchmark Details}

\paragraph{Evaluation Metrics.} During evaluation, the visual quality is quantified through PSNR, SSIM~\cite{wang2004image}, LPIPS~\cite{Johnson2016Perceptual}, FID~\cite{heusel2017gans}, and FVD~\cite{unterthiner18}, and the trajectory fidelity is quantified by Average Displacement Error (ADE) and Final Displacement Error (FDE) upon the trajectory pair $\{\mathbf{y}_t, \hat{\mathbf{y}}_t\}$ estimated via MonST3R:
\begin{align}
\text{ADE} = \frac{1}{T} \sum_{t=1}^{T} \| \mathbf{y}_t - \hat{\mathbf{y}}_t \|_2 , \text{FDE} = \| \mathbf{y}_T - \hat{\mathbf{y}}_T \|_2 ~,
\end{align}
where $\mathbf{y}_t$ denotes ground truth poses and $\hat{\mathbf{y}}_t$ predicted positions, and T is the total
number of frames. 

\section{Additional Results}
\label{sec:append_results}

\subsection{Generalization to Unseen Trajectory}
\begin{figure*}[h]
  \centering
   \includegraphics[width=0.9\linewidth]{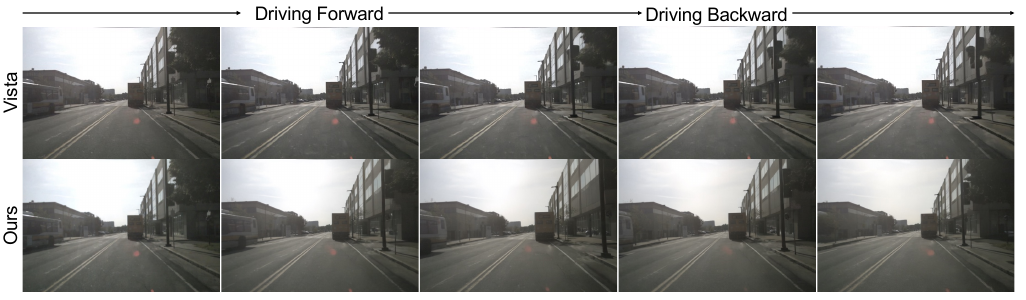}
   \caption{Our model can faithfully follow given trajectory and predict consistent future, even when such trajectory is out of training distribution. Yet previous work fail to follow the given action.}
   \label{fig:qual_reverse}
\end{figure*}

n Section~\ref{sec:result_trajectory}, we highlight the superior trajectory-following capabilities of GeoDrive compared to baseline methods. GeoDrive's controllability naturally extends to unseen trajectories, such as shifted and reverse trajectories (Section~\ref{sec:result_novelview}). Figure~\ref{fig:qual_reverse} illustrates this, where both Vista and GeoDrive were provided with a reverse trajectory. Unlike Vista, which only moves forward, GeoDrive accurately follows the instructions, moving first forward and then backward. This discrepancy arises because reverse trajectories are rare in the training data, causing Vista's implicit action control to struggle with generalization. In contrast, GeoDrive's design, which utilizes renderings as conditions, allows it to generalize effectively to any trajectory.

\end{document}